\documentclass{article}
\usepackage{ijcai26}

\usepackage{times}
\usepackage{soul}
\usepackage{url}
\usepackage[hidelinks]{hyperref}
\usepackage[utf8]{inputenc}
\usepackage[small]{caption}
\usepackage{graphicx}
\usepackage{amsmath}
\usepackage{amsthm}
\usepackage{booktabs}
\usepackage{algorithm}
\usepackage{algorithmic}
\usepackage[switch]{lineno}
\usepackage{multicol}
\usepackage{multirow}
\usepackage{xcolor}
\usepackage{amssymb}
\usepackage{enumitem}
\usepackage{natbib}

\title{Enhancing Cross-Problem Vehicle Routing via Federated Learning}

\author{
Xiangchi Meng$^{1*}$\and
Jianan Zhou$^{2}$\thanks{Equal contribution.}\and
Jie Gao$^3$\and
Yifan Lu$^1$\\
Yaoxin Wu$^4$\and
Gonglin Yuan$^1$\and
Yaqing Hou$^1$\\
\affiliations
\vspace{1mm}
$^1$Dalian University of Technology\\
$^2$Nanyang Technological University\\
$^3$Delft University of Technology\\
$^4$Eindhoven University of Technology
}
\begin{document}

\maketitle

% \begin{abstract}
% Vehicle routing problems constitute a core optimization challenge in modern logistics and supply chain management. In recent years, while Neural Combinatorial Optimization has demonstrated efficiency superior to traditional algorithms, existing Cross-Problem Learning paradigms often encounter bottlenecks such as performance degradation and generalization decay when transferring from simple tasks to heterogeneous scenarios involving multiple complex constraints. To address these challenges, this paper proposes an innovative ``Multi-Task Pre-training — Single-Task Fine-tuning'' framework with Federated Learning mechanisms to enhance Cross-Problem Learning paradigms. This strategy aims to leverage the shared common knowledge of the federated global model, achieving efficient cross-task knowledge sharing and transfer without altering the single-task fine-tuning setting. This enables local solvers to effectively retain memory of shared common knowledge while adapting to specific constraints. Experimental results demonstrate that our framework not only mitigates transfer performance degradation under complex constraints, but also enhances the unseen problems generalization capability and robustness of neural solvers in multi-task scenarios while protecting data privacy.

% \end{abstract}

\begin{abstract}
     Vehicle routing problems (VRPs) constitute a core optimization challenge in modern logistics and supply chain management. The recent neural combinatorial optimization (NCO) has demonstrated superior efficiency over some traditional algorithms. While serving as a primary NCO approach for solving general VRPs, current cross-problem learning paradigms are still subject to performance degradation and generalizability decay, when transferring from simple VRP variants to those involving different and complex constraints. To strengthen the paradigms, this paper offers an innovative ``Multi-problem Pre-train, then Single-problem Fine-tune'' framework with Federated Learning (MPSF-FL). This framework exploits the common knowledge of a federated global model to foster efficient cross-problem knowledge sharing and transfer among local models for single-problem fine-tuning. In this way, local models effectively retain common VRP knowledge from up-to-date global model, while being efficiently adapted to downstream VRPs with heterogeneous complex constraints. Experimental results demonstrate that our framework not only enhances the performance in diverse VRPs, but also improves the generalizability in unseen problems.
% and the robustness of neural solvers in multi-task scenarios while protecting data privacy.

\end{abstract}

\section{Introduction}

Vehicle Routing Problems (VRPs) serve as a cornerstone of combinatorial optimization, exerting a profound influence on the efficiency of modern logistics, e-commerce, and supply chain management ~\citep{cattaruzza2017vehicle,konstantakopoulos2022vehicle}. In recent years, Neural Combinatorial Optimization (NCO) has emerged as a transformative paradigm for addressing VRPs. Deep learning-based neural solvers are capable of generating near-optimal solutions at a computational cost significantly lower than that of traditional heuristics or exact algorithms, while possessing competitive solutions ~\citep{kwon2020pomo,li2021learning,kim2022sym,luo2023neural,gao2024towards,ma2025coexpander}.

% \begin{figure}[htbp]
%   \centering
%   \includegraphics[width=0.45\textwidth]{figure/performance_drop.png}
%   \caption{Performance comparison of Cross-Problem Learning from different initialization method on OVRPB, VRPLTW, OVRPBTW, OVRPBLTW. These VRP variants will be discussed in Section \ref{sec:vrp_variants}. To ensure fairness, identical model architectures are employed for comparison.}

%   \label{fig:performance_drop}
% \end{figure}

% \begin{figure}[htbp]
%   \centering
%   \includegraphics[width=0.45\textwidth]{figure/forgetting.png}
%   \caption{Illustration of generalization performance decay when pre-trained model trained by Cross-Problem Learning on OVRPB. The generalization performance of the model fine-tuned by CPL exhibits varying degrees of decline across many tasks. 
%   % \textbf{Blue bars} represent models pre-trained on a mixture of multiple VRP tasks; \textbf{Red bars} represent models cross-problem fine-tuning on OVRPB. Traditional cross-problem fine-tuning on specific variants leads to a loss of global shared common knowledge.
%   }
%   \label{fig:forgetting}
% \end{figure}

Notwithstanding these remarkable achievements, existing methodologies typically train neural solvers for specific VRPs in isolation, necessitating training a new model from scratch for every problem type. Consequently, neural solvers learned for one VRP are often discarded when confronted with new variants, leading to scarce transferable knowledge and unnecessarily high training costs. In response, a recent Cross-Problem Learning (CPL) paradigm has been proposed to facilitate parameter transfer using generic components (a.k.a. the backbone) of Transformer networks~\citep{lin2024cross}. The backbone trained on the Traveling Salesman Problem (TSP) captures transferable knowledge that, when fine-tuned on other VRPs, yields improved convergence and performance relative to training from scratch.
% and fosters the training of neural solvers for other VRPs.

\begin{figure}[t!]
  \centering
  \includegraphics[width=1.0\linewidth]{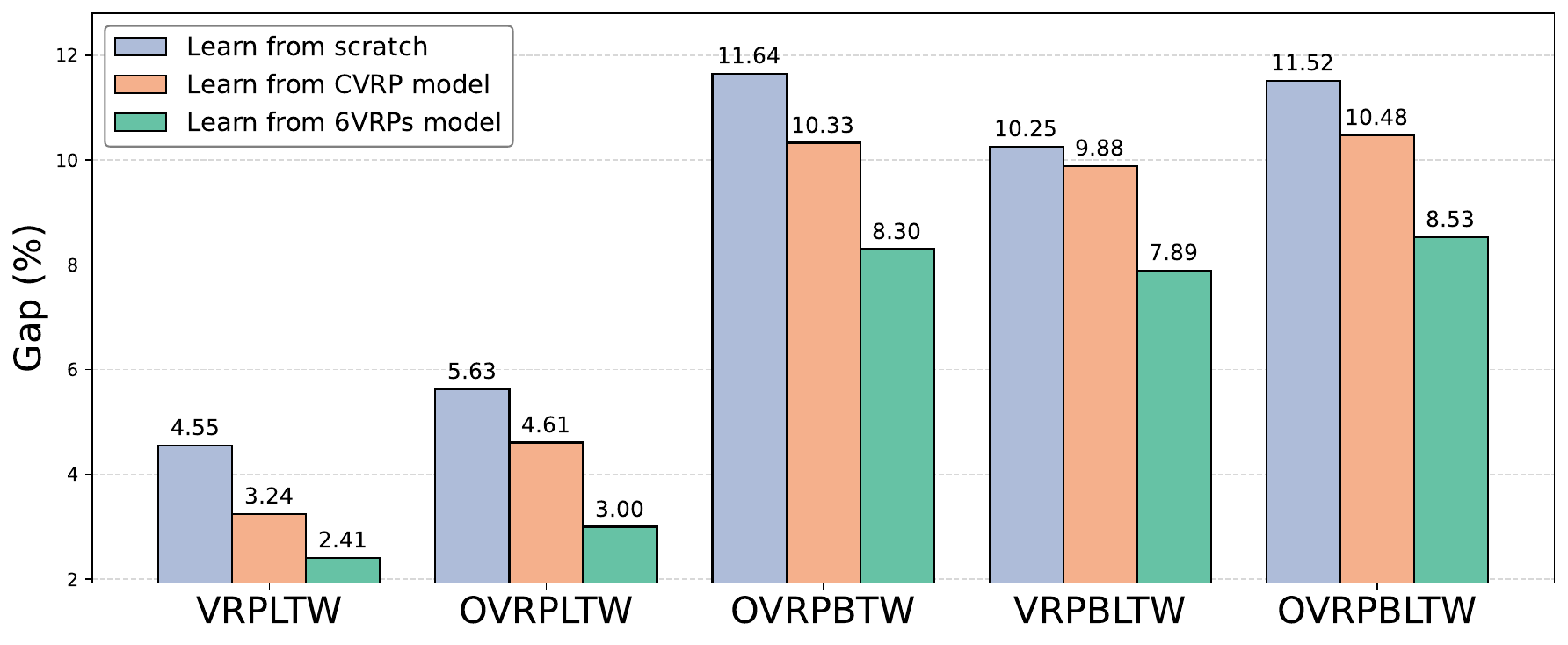}
  \caption{Performance comparison of CPL. Neural solvers (with identical neural architectures) trained on CVRP and 6 VRPs are fine-tuned on VRPLTW, OVRPLTW, OVRPBTW, VRPBLTW, OVRPBLTW, respectively. These VRP variants are described in Section \ref{sec:vrp_variants}.}
  \label{fig:performance_drop}
\end{figure}

\begin{figure}[t!]
  \centering
  \includegraphics[width=\linewidth]{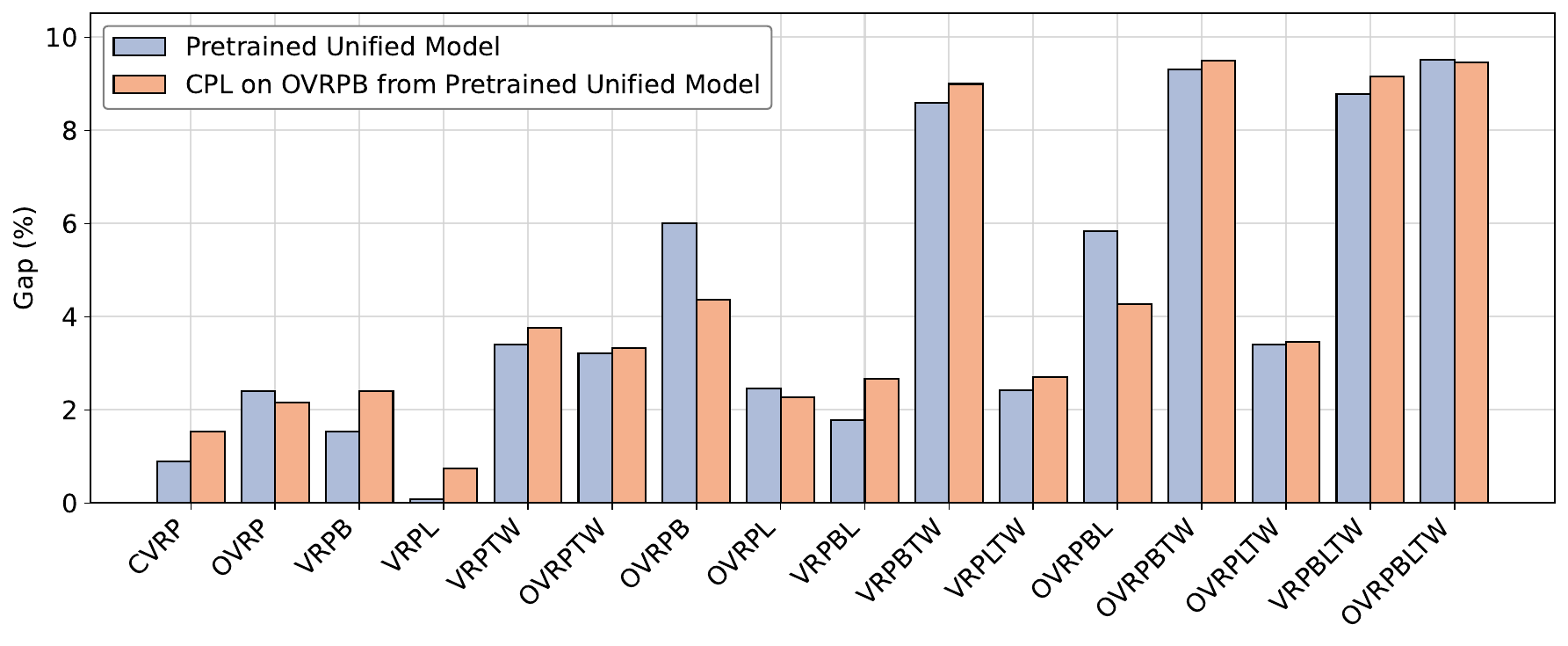}
  \caption{Generalization performance decay when a pre-trained unified model is fine-tuned by CPL on OVRPB.}
  \label{fig:forgetting}
\end{figure}

However, the aforementioned CPL postulates knowledge transfer from a single simple VRP (e.g. TSP, CVRP), thus constraining its transferability to similar problem variants and hindering its few-shot performance. More specifically, the efficacy of CPL deteriorates significantly when transferring from simple VRPs to those with multiple constraints. As illustrated in Figure~\ref{fig:performance_drop}, when a neural solver is transferred from CVRP to more complex variants such as VRPLTW, OVRPLTW, OVRPBTW, VRPBLTW and OVRPBLTW via CPL, its performance is similar to that of solvers trained from scratch with randomly initialized parameters. In contrast, neural solvers trained on 6 different VRPs consistently yield much lower objective gaps. Hence some unified neural solvers trained on multiple VRPs have applied fine-tuning to enhance their few-shot results~\citep{zhou2024mvmoe,berto2025routefinder}.
% As shown in Figure~\ref{fig:performance_drop}, when transferring a neural solver from CVRP to more complex OVRPB, VRPLTW, OVRPBTW and OVRPBLTW via CPL, the solver exhibits similar performance to those trained from scratch with randomly initialized parameters. 
% This demonstrates that the traditional CPL paradigm faces certain limitation in such scenarios. 
% This phenomenon is attributed to the inability of representations learned via single-problem training to maintain robustness under drastic variations in problem constraints. 
% Moreover, 
% when fine-tuning on VRP variants,
% traditional CPL paradigms may overfit to variant-specific constraints, thereby losing common knowledge in the backbone and degrading overall generalization performance.
On the other hand, fine-tuning on VRP variants under traditional CPL paradigms may risk overfitting to variant-specific constraints, which erodes the backbone’s shared knowledge and impairs generalization.
% catastrophic forgetting of global common knowledge due to overfitting to local constraints, thereby degrading generalization performance.
Figure~\ref{fig:forgetting} illustrates that after fine-tuning a (pre-trained) unified neural solver on OVRPB, its ability to generalize to other VRP variants deteriorates, leading to inferior zero-shot performance compared to the original unified solver.
% after a (pre-trained) unified neural solver is fine-tuned on OVRPB, it exhibits inferior generalizable performance on other VRP variants, indicating underperformance over the original unified solver in a zero-shot manner. 
% This oversight in traditional CPL paradigms regarding single-task cross-problem fine-tuning severely limits the practical utility of such models.

To further strengthen CPL of unified neural solvers, 
% this work aims to overcome the representation limitations to effectively share common knowledge, thereby enhancing the performance of CPL in solving  VRPs with complex, heterogeneous constraints. 
this work aims to leverage the common knowledge from a global model as implicit guidance to simultaneously enhance the fine-tuning of neural solvers for different VRPs. As such, each neural solver can gain better performance for the corresponding VRP and retain good generalizability across other VRP variants. We propose a ``Multi-problem Pre-train, then Single-problem Fine-tune'' with Federated Learning (MPSF-FL), a novel CPL framework that relies on a federated global model to enhance the fine-tuning of local neural solvers, even without explicit data sharing between clients (i.e. local VRPs). 
% which innovatively applies federated learning to address the deficiencies of traditional CPL paradigm in knowledge transfer among complex constraints. 
The ties-merging strategy is further adopted to addresses the parameter confliction caused by multi-task aggregation, thereby comprehensively raising the performance of local neural solvers.
% Moreover, we investigate \textcolor{red}{generalization methodologies and collaborative training modes} for multi-task neural solvers, extending the applicability of CPL. 
Our contributions are summarized as follows:

\begin{enumerate}[label=\arabic*)]
    % \item \textbf{Revealing and mitigating CPL transfer degradation to complex constraints:} We identify performance bottlenecks when transferring backbones trained on simple tasks to those with complex constraints. By employing an architecture with superior representation capabilities and utilizing multi-task learning to cover VRPs with diverse constraints, we construct a robust pre-trained model for CPL. This enables preliminary modeling of heterogeneous tasks, thereby enhancing the efficacy and robustness of the CPL paradigm.

    % \item \textbf{Enhancing CPL paradigm via Federated Learning:} Addressing the prevalent decline in generalization performance during new problem adaptation, we introduce a innovative CPL framework with federated learning mechanism while preserving the single-task fine-tuning setting.  This enables the model to acquire knowledge from multiple other problems even use CPL locally, significantly bolstering the local solver's generalization performance on unseen variants.

    \item We propose MPSF-FL, a federated learning-based CPL framework, by which local neural solvers are simultaneously fine-tuned by employing knowledge from a federated global model updated via aggregating local solvers. 
    % This CPL framework significantly bolsters the local solvers' performance in corresponding VRPs and generalizability in unseen variants.    

    \item We adopt a ties-merging strategy to facilitate the aggregation of the federated global model , and therefore further enhances the performance of local neural solvers.

    % \item We apply the proposed MPSF-FL framework on XXXX VRPs effectively mitigates aggregation performance degradation caused by Non-IID data in multi-task federated scenarios~\citep{li2020federated,collins2021exploiting}. Furthermore, it demonstrates robustness in offline settings simulating real-world applications, all while safeguarding data privacy.
    \item We apply MPSF-FL to two unified neural solvers (MVMoE and CaDA), which demonstrates improved performance and generalizability in 10 diverse VRPs. It shows advantage over traditional CPL fine-tuning for each local VRP and the fine-tuning on all VRPs.
    % Fine-tuning with very limited local datasets verifies the advantage of MPSF-FL, when involving practical data scarcity and privacy.
\end{enumerate}

\section{Related Work}

\subsection{Neural VRP solvers}

A predominant NCO paradigm for solving VRPs is constructive neural solvers that focus on learning solution generation policies in an end-to-end manner. \cite{vinyals2015pointer} pioneered the pointer network to construct TSP solutions via an autoregressive way. \cite{bello*2017neural} introduced reinforcement learning (RL) to explore superior approximate solutions for TSP. \cite{nazari2018reinforcement} extended this method to CVRP. Afterward, \cite{kool2018attention} proposed the attention model (AM), adapting the Transformer architecture to solve various VRP variants independently. Exploiting solution symmetries, \cite{kwon2020pomo} proposed policy optimization with multiple optima (POMO), which improved AM performance on TSP and CVRP. Most following constructive solvers have been developed upon  POMO~\citep{kwon2021matrix,grinsztajn2023winner,bi2024learning,hottung2025polynet,huang2025rethinking,liao2025bopo}. However, these methods require training a model from scratch for each specific problem, leading to limited knowledge transfer and unnecessarily high training costs. \cite{lin2024cross} proposed the CPL paradigm to reuse knowledge captured by a backbone trained on TSP, but overlooked the fact that unified models exhibit stronger transfer learning capabilities, as discussed in the Introduction.
% substantial computational overhead and memory costs~\citep{lin2024cross}.

% However, these methods face a critical limitation when encountering new problems. They typically require training the model from scratch for each specific problem, resulting in substantial computational overhead and memory costs~\citep{lin2024cross}.

\subsection{Unified Models for VRPs}

A unified model aims to solve diverse problems by jointly learning multiple VRP variants, thereby avoiding unnecessary separate training efforts. \cite{zhou2024mvmoe} proposed MVMoE, a multi-task model with mixture-of-experts that improves solution accuracy on heterogeneous tasks through the decoupling and collaboration of expert modules. \cite{goh2025shieldmultitaskmultidistributionvehicle} proposed SHIELD, a model integrating sparsity and hierarchy principles on top of MVMoE to boost both efficiency and generalization. \cite{berto2025routefinder} utilized Transformer encoder and global attribute embeddings to enhance VRP representation capabilities. \cite{pan2025moses} introduced a state-decomposable Markov decision process (SDMDP) alongside a mixture-of-specialized-experts solver (MoSES).  Differently, \cite{li2025cada} introduced the constraint-aware dual-attention (CaDA) model for overcoming the limitations of constraint-agnostic approaches. Despite their promising performance, unified neural solvers still exhibit limited effectiveness when fine-tuned on an unseen VRP variant using CPL~\citep{zhou2024federated,li2025cada}. This paper aims to enhance CPL performance via federated learning that simultaneously trains neural solvers across multiple VRP variants.

% However, applying these methods in local sides presents a practical challenge. The scarcity of data for specific problem types leads to the lack of comprehensive data hinders the feasibility of local multi-task learning, making it difficult to obtain a unified model that has robust performance across all tasks.

\subsection{Federated Learning}

To address data privacy concerns, \cite{mcmahan2017communication} established the federated learning paradigm, enabling decentralized clients to collaboratively train a global model without sharing their own data. However, in real-world, data distributions across clients are often Non-IID. To address this challenge, \cite{smith2017federated} introduced the federated multi-task learning (FMTL) framework, which treats each client as a distinct task. \cite{MLSYS2020_1f5fe839} proposed adding a proximal term to the local objective function to constrain the deviation of local updates from the global model, thereby addressing both statistical and system heterogeneity. Follow-up works mainly address Non-IID client data distributions, enabling heterogeneous individual clients to leverage global knowledge for meeting specific task requirements \citep{fallah2020personalized,sattler2020clustered,morafah2023flis,wang2023fedftha,zhou2024federated,wang2024flora,hu2025fft}.

\section{Preliminaries}

\subsection{VRP Formulation and Variants}
\label{sec:vrp_variants}

% A generic Vehicle Routing Problem instance is defined on a fully connected graph $\mathcal{G} = (\mathcal{V}, \mathcal{E})$, where the node set $\mathcal{V} = \{v_0, v_1, \ldots, v_N\}$ includes a single depot $v_0$ and $N$ customer nodes. The edge set is defined as $\mathcal{E} = \mathcal{V} \times \mathcal{V}$. Each node $v_i \in \mathcal{V}$ is described by a coordinate vector $\vec{X}_i \sim U(0,1)^2$ and a specific set of attributes $A_i$. Transportation costs are given by a cost matrix $\mathbf{D} = \{ d_{i,j} \}$ corresponding to Euclidean distances.

% For the Capacitated Vehicle Routing Problem (CVRP), the depot attribute set $A_0 = \emptyset$, while each customer node $v_i$ is associated with a non-negative demand $\delta_i$ (i.e., $A_i = \{\delta_i\}$). The fleet consists of $K$ homogeneous vehicles, each with a maximum load capacity of $C$. The solution is represented as a set of vehicle routes $\boldsymbol{\pi} = \{\boldsymbol{\pi}^1, \ldots, \boldsymbol{\pi}^K\}$. Here, each sub-route $\boldsymbol{\pi}^k = (\pi^k_1, \ldots, \pi^k_{n_k})$ starts and ends at the depot $v_0$ ($\pi^k_1 = \pi^k_{n_k} = 0$), serves a subset of customers, and ensures the total load does not exceed the capacity limit $C$ (i.e., $\sum_{v_j \in \boldsymbol{\pi}^k} \delta_j \le C$). The objective is to find a solution $\boldsymbol{\pi}^*$ that minimizes the total trajectory distance.

A basic capacitated vehicle routing problem (CVRP) is defined on a graph $\mathcal{G} = (\mathcal{V}, \mathcal{E})$, comprising a depot node $v_0$ and $n$ customer nodes $\{v_1, \ldots, v_n\}$. Each node is associated with a Euclidean coordinate vector, and each customer $v_i$ features a demand $\delta_i$. The objective is to minimize the total trajectory distance of all vehicles with capacity $c$. A feasible solution $\boldsymbol{\pi}$ consists of $\mathbf{K}$ routes that start and end at the depot, visit each customer exactly once while satisfying the capacity constraint such that the total demand within each route of vehicle $\mathbf{k}$ does not exceed $c$, i.e., $\sum_{v_j \in \boldsymbol{\pi}^\mathbf{k}} \delta_j \le c$. Based on CVRP, this paper explores 4 additional constraints commonly involved in recent unified models~\citep{liu2024multi,zhou2024mvmoe,li2025cada}. Combinations of these constraints lead to 6 simple VRP variants and 10 complex variants. The constraints are defined as follows:

\begin{itemize}
    \item \textbf{Open VRP (O-VRP):} Unlike CVRP, vehicles do not return to the depot after serving the last customer on the route, upon which their routes terminate immediately.
    
    \item \textbf{VRP with Backhauls (VRP-B):} Customer nodes are classified as linehaul customers, with demand $\delta_i > 0$, requiring deliveries from the depot, and backhaul customers, with demand $\delta_i < 0$, requiring the collection of $|\delta_i|$ goods to be returned to the depot.
    
    % Customer nodes are divided into linehaul customers with demand $\delta_i > 0$, requiring goods to be delivered from the depot to their locations, and backhaul customers with demand $\delta_i < 0$, requiring the vehicle to collect $|\delta_i|$ goods from their locations and transport them back to the depot.
    
    \item \textbf{VRP with Duration Limit (VRP-L):} The total trajectory distance of every vehicle must adhere to a preset maximum limit $L$.
    
    \item \textbf{VRP with Time Windows (VRP-TW):} Arrival at each customer $v_i$ must be within a time window $[e_i, l_i]$, where $e_i$ and $l_i$ are the earliest and latest start time. Besides, serving a customer requires a duration $s_i$.
\end{itemize}

\subsection{Basics for Unified Neural Solvers}
Constructing VRP solutions follows a Markov decision process (MDP). In the NCO field, the policy is typically parameterized by a Transformer with parameters $\theta$. During solution construction, it infers the probability distribution over nodes based on the current state, including static features of the VRP instance and dynamic features of the current partial solution. The probability of generating a route $\pi$ is formulated by:
\begin{equation}
    p_{\theta}(\pi|\mathcal{G}) = \prod_{t=1}^{T} p_{\theta} (\pi_{t}|\mathcal{G}, \pi_{<t})
\end{equation}
where $\pi_t$ denotes the node selected at time step $t$, and $\pi_{<t}$ denotes the partial route constructed up to time step $t$. The REINFORCE algorithm~\citep{williams1992simple} is often employed to optimize the policy, with gradients defined below:
\begin{equation}
\label{local_loss}
    \nabla_{\theta} \mathcal{L}(\theta|\mathcal{G}) = \mathbb{E}_{p_{\theta}(\pi|\mathcal{G})} \left[ (L(\pi)-b(\mathcal{G})) \nabla_{\theta}\log p_{\theta}(\pi|\mathcal{G}) \right]
\end{equation}
where $L(\pi)$ represents the distance of the generated route $\pi$ for instance $\mathcal{G}$, and $b(\mathcal{G})$ is a baseline function used to reduce the variance of the gradient estimation. Current unified neural solvers employ the above basics to train the Transformer across multiple VRPs. E.g., MVMoE trains a unified neural solver on six VRPs by extending the Transformer with mixture-of-experts~\citep{zhou2024federated}. 

\section{Methodology}

\begin{figure*}[t]
    \centering
    \includegraphics[width=0.86\textwidth]{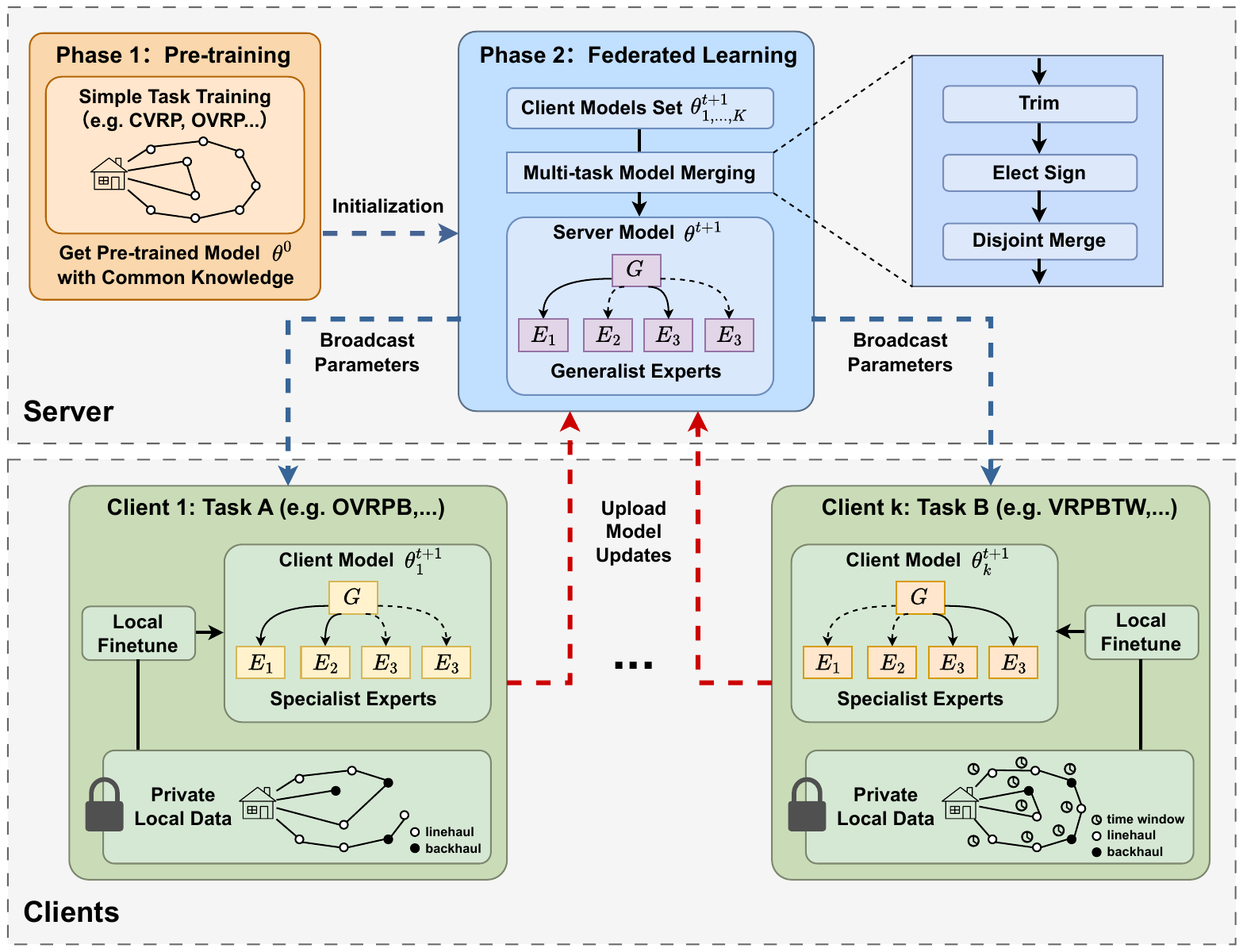}
    \caption{Our ``Multi-problem Pre-train, then Single-problem Fine-tune'' framework with Federated Learning (MPSF-FL) for VRPs, taking MVMoE as a model example. The $G$ and $\{E_1, E_2, E_3, E_4\}$ in the figure represent the gating network and expert network in the mixture-of-expert moudle of MVMoE, which handles different problem features. 
    % Details of the mixture-of-expert moudle are presented in Appendix \ref{app:mvmoe}.
    }
    \label{fig:framework}
\end{figure*}

\subsection{Overall MPSF-FL Framework}

% Our MPSF-FL aims to address the adaptation difficulties and generalization degradation encountered by traditional CPL paradigms when dealing with complex VRPs scenarios. 
As illustrated in Figure \ref{fig:framework}, the ``Multi-problem Pre-train, then Single-problem Fine-tune'' framework with Federated Learning (MPSF-FL) represents a distributed collaborative system composed of a central server and multiple clients. Specifically, $N$ local clients, indexed by $\mathcal{N} = \{1, \dots, N\}$, represent different local VRP variants (e.g., arising from practical delivery scenarios) participating in the federated learning. Each client $i$ possesses a private dataset local $\mathcal{D}_i$ following a specific distribution for a local VRP, with a data size $|\mathcal{D}_i|$. A unified neural solver is distributed to each client for fine-tuning it on local VRPs, respectively. After fine-tuning, the central server, responsible for aggregating local neural solvers and thus extracting common knowledge to a global model, aims to collaboratively minimize a global loss function:
\begin{equation}
    \min_{\theta} F(\theta) = \sum_{i=1}^N p_i \mathcal{L}_i(\theta)
\end{equation}
where $\theta$ signifies parameters of the global model, $p_i$ indicating the importance weight for client $i$, typically set to $\frac{|\mathcal{D}_i|}{\sum_{j=1}^N |\mathcal{D}_j|}$, 
% and $\mathcal{L}_k(\theta) = \mathbb{E}_{(x,y) \sim \mathcal{D}_k} [\ell(\theta; x, y)]$
and $\mathcal{L}_i(\theta)$ is the loss function of client $i$ used for fine-tuning on its local dataset $\mathcal{D}_i$. By distributing global model, collecting local gradients and aggregating parameters of local solvers, the server optimizes the objective to acquire cross-problem general knowledge. More specifically, the MPSF-FL process is divided into two phases: \textbf{Multi-problem Pre-train} and \textbf{Single-problem Fine-tune}, elaborated below.

\subsection{Multi-problem Pre-train}

% As discussed in Introduction, to ensure the representation bottleneck of traditional CPL paradigm in complex constraints learning, we first improves it by introducing unified multi-task model, a Multi-Task Vehicle Routing Solver based on Mixture of Experts (MVMoE)~\citep{zhou2024mvmoe}, with parameters denoted as $\theta^0$. The detailed structure of this model is presented in Appendix \ref{app:mvmoe}.

As discussed in Introduction, the traditional CPL postulates knowledge
transfer from a single simple VRP (TSP or CVRP)~\citep{lin2024cross}, which performs inferior to CPL using a unified neural solver. Therefore, we instantiate the global model by a unified neural solver such as MVMoE~\citep{zhou2024mvmoe} with parameters $\theta^0$. The learning objective of this phase is to capture general VRP representations by training the global model on $M$ simple  VRPs $\mathcal{T}_{pre}=\{T_1, \ldots, T_M\}$.
% with different attribute distributions. 
The server centrally minimizes the expected loss across the VRPs:
\begin{equation}
    \theta^0 = \arg \min_{\theta} \sum_{m=1}^{M} \mathbb{E}_{\mathcal{G} \sim \mathcal{P}_m} [\mathcal{L}(\theta|\mathcal{G})]
    \label{eq:pretrain}
\end{equation}
where $\mathcal{P}_m$ denotes the instance distribution of the $m$-th VRP. Given a unified neural solver, we employ the REINFORCE algorithm to minimize the loss and simultaneously involves associated techniques to enhance the performance, such as the balancing loss for the mixture of experts in MVMoE. 
% Details of MVMoE pre-training are presented in Appendix \ref{sec:pretrain_mvmoe}.
% Thus, the total loss is defined as $\mathcal{L} = \mathcal{L}_a + \alpha \mathcal{L}_b$, where $\alpha$ is a hyperparameter controlling the intensity of the balancing loss.
After multi-problem pre-training, the global model is updated to obtain a unified solver capable of solving the trained VRPs as well as unseen VRPs with combinations of constraints drawn from the trained problems. The pre-trained global model potentially captures common knowledge such as local structure and route connectivity, offering an effective inductive bias for subsequent  fine-tuning on VRPs of clients.

% It is worth noting that although the model structure itself is homogeneous, through joint training on diverse tasks, its internal MoE module functionally behaves as a generalist expert. It learns to handle basic node neighborhood relations and path connectivity, thereby providing a robust inductive bias for subsequent transfer to complex constraint combinations.

\begin{algorithm}[tb]
    \caption{Multi-Task Federated Learning on VRPs}
    \label{alg:fl}
    {\raggedright
    \textbf{Input}: Pre-trained server model $\theta^0$; Set of $N$ clients with private datasets $\{\mathcal{D}_1, \ldots, \mathcal{D}_N\}$ \\
    \textbf{Parameter}: Client selection ratio $C$; Local training epochs $E$; Local learning rate $\eta$; Balancing loss parameter $\alpha$; Total communication rounds $T$ \\
    \textbf{Output}: Set of client models $\{\theta_1^T, \ldots, \theta_N^T\}$ 
    \par}
\begin{algorithmic}[1]
\STATE Initialize global model with pre-trained parameters $\theta^0$.
\FOR{$t = 0, 1, \ldots, T-1$}
    \STATE \textbf{Server:}
    \STATE $K \leftarrow \max(C \cdot N, 1)$
    \STATE $S_t \leftarrow$ Randomly select $K$ clients
    \STATE Broadcast global model parameters $\theta^t$ to all selected clients $S_t$
    \FOR{each client $k \in S_t$ \textbf{in parallel}}
        \STATE \textbf{Client $k$:}
        \STATE Initialize local model $\theta_k^t \leftarrow \theta^t$
        \FOR{$e = 1, \ldots, E$}
            \STATE Compute gradient $g \leftarrow \nabla_{\theta} \mathcal{L}(\theta_k^t; \mathcal{D}_k)$
            % \IF{$t = 1$}
            %     \STATE $\mathcal{L} = \mathcal{L}_a + \alpha \mathcal{L}_b$
            % \ELSE
            %     \STATE $\mathcal{L} = \mathcal{L}_c + \alpha \mathcal{L}_b$
            % \ENDIF
            \STATE $\theta_k^{t, e+1} \leftarrow \theta_k^{t, e} - \eta g$
        \ENDFOR
        \STATE Send $\theta_k^{t+1}$ to Server
    \ENDFOR
    \STATE \textbf{Server:}
    \STATE $\theta^{t+1} \leftarrow \text{Aggregate}(\{\theta_k^{t+1}\}_{k \in S_t})$
\ENDFOR
\RETURN $\{\theta_1^T, \ldots, \theta_N^T\}$
\end{algorithmic}
\end{algorithm}

% \subsection{Federated Learning on Complex Problems}
\subsection{Single-problem Fine-tune}

% The federated learning process serves not only as a means of distributed training but also as a critical implicit regularizer. 
This phase serves not only as a mechanism for distributed training but also as a critical implicit regularizer. Its objective is to coordinate the fine-tuning on respective VRPs of clients, while protecting data privacy, while preventing generalizability decay and potential overfitting to every specific VRP.
% caused by local cross-problem fine-tuning.  
% while protecting data privacy, preventing generalization decay and potential overfitting caused by local cross-problem fine-tuning.
The training process of this phase consists of three main steps:

\begin{enumerate}[label=\arabic*)]
    \item \textbf{Global Initialization:} The central server distributes the global model's parameters $\theta^0$ to all $N$ clients, which share common knowledge obtained from pre-training phase.
    % , to all $K$ clients $\mathcal{C}$. 
    As such, all clients start their fine-tuning from a well pre-trained backbone.
    % rather than random initialization.
    
    \item \textbf{Local CPL:} In each communication round $t$, server select $K$ clients to obtain the set of clients $S_t$ participating in this round. Then the selected clients $S_t$ receive the global parameters $\theta^t$ from the server. 
    % To adapt to specific local VRP constraints, 
    Then, each client $k$ performs fine-tuning on its local VRPs with private dataset $\mathcal{D}_k$ for a total of $E$ epochs:
    \begin{equation}
        \theta_{k}^{t,e+1} \leftarrow \theta_{k}^{t,e} - \eta \nabla \mathcal{L}_k(\theta_{k}^{t,e}|\mathcal{D}_k)
        \label{eq:local_update}
    \end{equation}
    where $e$ is the current epoch, $\eta$ is the local learning rate, and $\mathcal{L}_k$ denotes the loss function for the local VRP. 
    % defined as $\mathcal{L}$ in Equation \ref{mvmoe_train_loss} (Appendix \ref{sec:pretrain_mvmoe}).
    % In this framework, the local update builds upon the standard REINFORCE algorithm by utilizing the client's historical model parameters $\theta_{k}^{t-1}$ for inference. The inference result serves as an additional baseline for the loss function, thereby accelerating the convergence speed of the client's model parameter distribution towards the local problem after receiving the global model parameters. Therefore, except for the communication round $t=1$, the gradient approximation for the policy during local fine-tuning:
    
    % \begin{equation}
    % \begin{split}
    %     \nabla_{\theta} \mathcal{L}_c(\theta|\mathcal{G}) = \mathbb{E}_{p_{\theta}(\pi|\mathcal{G})} \Big[ & \Big( (r(\pi) - b(\mathcal{G})) + \\
    %     & \beta \cdot (r(\pi) - b(\mathcal{G}; \theta^{t-1})) \Big) \\
    %     & \times \nabla_{\theta} \log p_{\theta}(\pi|\mathcal{G}) \Big], \text{if } t > 1.
    % \end{split}
    %     \label{eq:gradient}
    % \end{equation}
    % where $\beta$ is a hyperparameter that controls the strength of the historical baseline.
    
    \item \textbf{Global Aggregation:} The server collects updated model parameters from all clients. An aggregation operator $\text{Agg}(\cdot)$ is performed to update the global model:
    \begin{equation}
        \theta^{t+1} \leftarrow \text{Agg}(\{\theta_{k}^{t+1}\}_{k\in S_t})
        \label{eq:aggregate}
    \end{equation}
    Given the parameter conflicts and redundancy issues that may arise from  simple federated averaging in multi-task federated learning scenarios ~\citep{yadav2023tiesmerging,yu2024language}, we  specialize $\text{Agg}(\cdot)$ for local neural solvers of VRPs in Section \ref{sec:merging}.
\end{enumerate}

The algorithm for this phase is presented in Algorithm~\ref{alg:fl}.

\subsection{Model Merging for Global Aggregation}
\label{sec:merging}

% In the aggregation of multi-task federated learning, simply employing traditional linear weighted averaging faces the challenges of \textbf{Parameter Conflicts} and \textbf{Parameter Redundancy}~\citep{yadav2023tiesmerging,yu2024language}.
% Simple model merging, such as averaging, would produce task-specific gradients that cancel each other out, causing that the global model to stagnate in these conflicting dimensions and weakening its adaptability to specific complex constraints, diluting the truly effective shared common knowledge.
Simple model merging strategies, such as parameter averaging, can produce task-specific gradients that cancel each other out, causing the global model to stagnate along conflicting dimensions. This, in turn, weakens its capability to adapt to complex, task-specific constraints and dilutes the truly effective shared knowledge. To address these issues, we extend ties-merging~\citep{yadav2023tiesmerging} strategy that relies on a core hypothesis: During multi-task fine-tuning, only a small fraction of parameters undergo significant and effective changes, and the update directions of different tasks on the same parameter dimension should remain consistent to avoid destructive interference. Following~\citep{ilharco2023editing}, we define the problem vector of client $k$ as $\tau_k = \theta_k^{t+1} - \theta^t$. We perform global aggregation by three sequential steps:

% \begin{itemize}
%     \item \textbf{Parameter Conflicts:} Since different clients optimize VRP variants with vastly different constraints, their gradient update directions may diverge drastically in the parameter space. Simple averaging would cause these task-specific gradients to cancel each other out causing the global model to stagnate in these conflicting dimensions and weakening its adaptability to specific complex constraints.
%     \item \textbf{Parameter Redundancy:} For a specific local task, often only a small subset of expert modules is significantly updated. Blindly averaging all parameters would introduce noise or invalid updates from large inactive regions into the global model~\cite{yu2024language}, diluting the truly effective shared common knowledge.
% \end{itemize}

% To address the aforementioned problems, we follow \textbf{Ties-Merging}~\citep{yadav2023tiesmerging} strategy. This method relies on a core hypothesis: during multi-task fine-tuning, only a small fraction of parameters undergo significant and effective changes, and the update directions of different tasks on the same parameter dimension should remain consistent to avoid destructive interference. Specifically, let the Task Vector~\citep{ilharco2023editing} of client $k$ be defined as $\tau_k = \theta_k^{t+1} - \theta^t$. The aggregation process performs through the following three sequential steps:

\begin{enumerate}[label=\arabic*)]
    % \item \textbf{Trim:} To address parameter redundancy, we first sparsify the update vector of each client. For vector $\tau_k$, the method retains only the top $\kappa\%$ parameters with the largest absolute values and sets the remaining parameters to zero, resulting in $\hat{\tau}_k$. This step not only filters out random noise during fine-tuning but also highlights the parameter adjustments most critical to the client's specific VRP task.
    \item \textbf{Trim:} To address parameter redundancy, we first sparsify the update vector of each client. For vector $\tau_k$, we retain the top $\kappa\%$ parameters with the largest absolute values and set the remaining parameters to zero, resulting in $\hat{\tau}_k$. In this regard, we filter out random noise during fine-tuning and underscore parameter adjustments that are most critical to the specific VRP of client $k$.    
    
    \item \textbf{Sign Voting:} Due to parameter conflicts, it is necessary to determine the dominant direction for updating the global model. Given each parameter dimension $j$, we aggregate directions of update vectors $\hat{\tau}_k^j$ from all participating clients. The final sign $\gamma^j$ is defined by: 
    % determined by the sign of the sum of all update vectors:
    \begin{equation}
        \gamma^j = \text{sgn}\left(\sum_{k \in S_t} \hat{\tau}_k^j\right)
    \end{equation}
    where $\text{sgn}(\cdot)$ is the sign function with $\text{sgn}(x) \cdot |x| = x$. This step ensures that the global update direction complies with the majority of local update directions.
    
    \item \textbf{Disjoint Merge:} 
    % After determining the dominant sign $\gamma^j$ for each parameter, 
    For each dimension $j$, we calculate the mean only for clients whose update direction is consistent with the selected sign $\gamma^j$. Updates contrary to the dominant direction are considered interference and set to zero. Formally, we define $\mathcal{A}^j = \{k \in S_t \mid \text{sgn}(\hat{\tau}_k^j) = \gamma^j\}$ and the merged vector dimension $j$ is calculated as:
    \begin{equation}
        \tau^j = \frac{1}{|\mathcal{A}^j|} \sum_{k \in \mathcal{A}^j} \hat{\tau}_k^j
    \end{equation}
\end{enumerate}

% These three steps not only achieve parameter sparsification but also eliminate sign conflicts between tasks.
Given the merged problem vector $\tau$, it is added back to the initial parameter values scaled by a hyperparameter $\lambda$, resulting in the final model parameters: $\theta^{t+1} = \theta^t + \lambda \cdot \tau$. 
% Through this algorithm, the framework continuously absorbs gradient information from different complex constraint tasks, ultimately achieving cross-problem generalization while safeguarding privacy.

\section{Experiments}

\begin{table*}[!t]
  % \vskip -0.05in
  \caption{Client-side performance comparison for MVMoE on 1K test instances. The best performance is highlighted in bold, and the second-best is underlined.}
  \label{mvmoe_client}
  % \vskip -0.1in
  \begin{centering}
  \begin{small}
  \renewcommand\arraystretch{1.05}  % 0.95
  \resizebox{1.0\textwidth}{!}{ 
  \begin{tabular}{ll|cccccccc|ll|cccccccc}
    \toprule
    \multicolumn{2}{c|}{\multirow{2}{*}{Method}} & \multicolumn{4}{c}{\textbf{$n=50$}} & \multicolumn{4}{c|}{$n=100$} & \multicolumn{2}{c|}{\multirow{2}{*}{Method}} &
    \multicolumn{4}{c}{\textbf{$n=50$}} & \multicolumn{4}{c}{$n=100$} \\
    \cmidrule(lr){3-6} \cmidrule(lr){7-10} \cmidrule(lr){13-16} \cmidrule(lr){17-20}
     & & Trained Obj. & Trained Gap & Unseen Obj. & Unseen Gap & Trained Obj. & Trained Gap & Unseen Obj. & Unseen Gap & & & Trained Obj. & Trained Gap & Unseen Obj. & Unseen Gap & Trained Obj. & Trained Gap & Unseen Obj. & Unseen Gap \\
     \midrule
     \multirow{6}*{\rotatebox{90}{OVRPB}} & OR-Tools & 5.764 & 0.332\% & 10.155 & 1.020\% & 8.522 & 1.852\% & 16.931 & 2.279\%  & \multirow{6}*{\rotatebox{90}{OVRPL}} & OR-Tools & 6.522 & 0.480\% & 10.071 & 1.003\% & 9.966 & 1.783\% & 16.770 & 2.286\%  \\ 
    & OR-Tools (x10) & 5.745 & * & 10.036 & * & 8.365 & * & 16.542 & *  & & OR-Tools (x10) & 6.490 & * & 9.953 & * & 9.790 & * & 16.383 & *  \\ 
    % & POMO-MTL-pretrain & 6.116 & 6.430\% & 10.665 & 6.065\% & 8.979 & 7.335\% & 17.532 & 6.109\%  & & POMO-MTL-pretrain & 6.668 & 2.734\% & 10.604 & 6.476\% & 10.126 & 3.441\% & 17.404 & 6.542\%  \\ 
    % & POMO-MTL-FL & 6.013 & 4.628\% & 10.655 & 5.838\% & 8.824 & 5.485\% & 17.545 & 6.018\% & & POMO-MTL-FL & 6.652 & 2.479\% & 10.592 & 6.189\% & 10.101 & 3.187\% & 17.414 & 6.385\%  \\ 
    & MVMoE-pretrain & 6.092 & 5.999\% &{\underline{10.640}} &{\underline{5.788\%}} & 8.959 & 7.088\% &{\underline{17.465}} &{\underline{5.706\%}} & & MVMoE-pretrain & 6.650 & 2.454\% &{\underline{10.578}} &{\underline{6.181}}\% & 10.097 & 3.148\% &{\underline{17.338}} &{\underline{6.143\%}}  \\ 
    & MVMoE-CPL &{\textbf{5.998}} &{\textbf{4.358\%}} & 10.656 & 5.833\% &{\textbf{8.788}} &{\textbf{5.048\%}} & 17.541 & 6.008\% & & MVMoE-CPL &{\textbf{6.633}} &{\textbf{2.192\%}} & {10.578} & {6.116\%} & {\textbf{10.061}} & {\textbf{2.780\%}} & 17.372 & 6.236\%  \\
    & MVMoE-MTF & 6.011  & 4.588\% & - & - & 8.802  & 5.211\% & - & - & & MVMoE-MTF & 6.640 & 2.295\% & - & - & {10.069}  &2.854\% & - & - \\
    & MVMoE-FL &{\underline{6.008}} &{\underline{4.545\%}} &{\textbf{10.623}} &{\textbf{5.528\%}} &{\underline{8.802}} &{\underline{5.211\%}} &{\textbf{17.475}} &{\textbf{5.572\%}} & & MVMoE-FL &{\underline{6.637}} &{\underline{2.251\%}} &{\textbf{10.560}} &{\textbf{5.922\%}} & 10.074 & 2.909\% &{\textbf{17.350}} &{\textbf{5.959\%}}  \\ 
    \midrule
    \multirow{6}*{\rotatebox{90}{VRPBL}} & OR-Tools & 8.131 & 1.254\% & 9.892 & 0.917\% & 12.095 & 2.586\% & 16.534 & 2.197\%  & \multirow{6}*{\rotatebox{90}{VRPBTW}} & OR-Tools & 15.053 & 1.857\% & 9.123 & 0.850\% & 26.217 & 2.858\% & 14.965 & 2.167\%  \\ 
    & OR-Tools (x10) & 8.029 & * & 9.782 & * & 11.790 & * & 16.161 & * & & OR-Tools (x10) & 14.771 & * & 9.033 & * & 25.496 & * & 14.638 & *  \\ 
    % & POMO-MTL-pretrain & 8.188 & 1.971\% & 10.435 & 6.561\% & 11.998 & 1.793\% & 17.196 & 6.725\% & & POMO-MTL-pretrain & 16.055 & 8.841\% & 9.561 & 5.797\% & 27.319 & 7.413\% & 15.494 & 6.100\%  \\
    % & POMO-MTL-FL & 8.187 & 1.964\% & 10.451 & 6.703\% & 12.005 & 1.855\% & 17.223 & 6.822\% &  & POMO-MTL-FL & 15.955 & 8.168\% & 9.577 & 5.944\% & 27.079 & 6.484\% & 15.527 & 6.280\%   \\ 
    & MVMoE-pretrain & 8.172 & 1.777\% &{\underline{10.409}} &{\underline{6.257\%}} &{\underline{11.945}} &{\underline{1.346\%}} &{\textbf{17.133}} &{\textbf{6.344\%}} & & MVMoE-pretrain & 16.022 & 8.600\% &{\underline{9.536}} &{\underline{5.499\%}} & 27.236 & 7.078\% &{\textbf{15.434}} &{\textbf{5.707\%}}  \\ 
    & MVMoE-CPL &{\textbf{8.168}} &{\textbf{1.732\%}} & 10.438 & 6.576\% &{\textbf{11.939}} &{\textbf{1.298\%}} & 17.174 & 6.638\% & & MVMoE-CPL &{\underline{15.921}} &{\underline{7.936\%}} & 9.600 & 6.406\% &{\underline{26.965}} &{\underline{6.030\%}} & 15.579 & 7.065\%  \\ 
    & MVMoE-MTF & 8.181 & 1.893\% & - & - & 11.971  & 1.563\% & - & - & & MVMoE-MTF &{\textbf{15.900}} &{\textbf{7.789\%}} & - & - &{\textbf{26.932}}  &{\textbf{5.892\%}} & - & -\\ 
    & MVMoE-FL &{\underline{8.172}} &{\underline{1.775\%}} &{\textbf{10.412}} &{\textbf{6.205\%}} & {11.957} & {1.446\%} &{\underline{17.167}} &{\underline{6.444\%}} & & MVMoE-FL & 15.925 & 7.948\% &{\textbf{9.533}} &{\textbf{5.406\%}} & 27.006 & 6.185\% &{\underline{15.447}} &{\underline{5.731\%}}  \\
    \midrule
    \multirow{6}*{\rotatebox{90}{VRPLTW}} & OR-Tools & 14.815 & 1.432\% & 9.150 & 0.898\% & 25.823 & 2.534\% & 15.009 & 2.203\% & \multirow{6}*{\rotatebox{90}{OVRPBL}} & OR-Tools & 5.771 & 0.549\% & 10.155 & 0.996\% & 8.555 & 2.459\% & 16.927 & 2.211\%  \\ 
    & OR-Tools (x10) & 14.598 & * & 9.052 & * & 25.195 & * & 14.672 & * & & OR-Tools (x10) & 5.739 & * & 10.036 & * & 8.348 & * & 16.543 & *  \\ 
    % & POMO-MTL-pretrain & 14.961 & 2.586\% & 9.683 & 6.492\% & 25.619 & 1.920\% & 15.683 & 6.710\% & & POMO-MTL-pretrain & 6.104 & 6.306\% & 10.667 & 6.079\% & 8.961 & 7.343\% & 17.534 & 6.108\%  \\ 
    % & POMO-MTL-FL & 14.967 & 2.631\% & 9.675 & 6.404\% & 25.605 & 1.869\% & 15.666 & 6.585\% & & POMO-MTL-FL & 6.008 & 4.655\% & 10.655 & 5.836\% & 8.809 & 5.522\% & 17.550 & 6.023\%   \\ 
    & MVMoE-pretrain & 14.937 & 2.421\% &{\underline{9.657}} &{\underline{6.185\%}} & 25.514 & 1.471\% &{\underline{15.625}} &{\underline{6.330\%}} & & MVMoE-pretrain & 6.076 & 5.843\% &{\underline{10.641}} &{\underline{5.805\%}} & 8.942 & 7.115\% &{\underline{17.467}} &{\underline{5.702\%}}  \\ 
    & MVMoE-CPL &{\textbf{14.924}} &{\textbf{2.328\%}} & 9.685 & 6.554\% &{\textbf{25.474}} &{\textbf{1.325\%}} & 15.730 & 7.168\% & & MVMoE-CPL &{\textbf{5.991}} &{\textbf{4.357\%}} & 10.658 & 5.853\% &{\textbf{8.767}} &{\textbf{5.020\%}} & 17.545 & 6.007\%  \\ 
    & MVMoE-MTF & 14.946  & 2.482\% & - & - & 25.525  & 1.526\% & - & - & & MVMoE-MTF & 6.000  & 4.517\% & - & - & 8.784  & 5.231\% & - & - \\
    & MVMoE-FL &{\underline{14.933}} &{\underline{2.405\%}} &{\textbf{9.636}} &{\textbf{5.935\%}} &{\underline{25.511}} &{\underline{1.464\%}} &{\textbf{15.600}} &{\textbf{6.098\%}} & & MVMoE-FL &{\underline{6.000}} &{\underline{4.517\%}} &{\textbf{10.625}} &{\textbf{5.534\%}} &{\underline{8.784}} &{\underline{5.230\%}} &{\textbf{17.477}} &{\textbf{5.565\%}}  \\ 
    \midrule
    \multirow{6}*{\rotatebox{90}{OVRPBTW}} & OR-Tools & 8.758 & 0.927\% & 9.823 & 0.954\% & 14.713 & 2.268\% & 16.243 & 2.233\% & \multirow{6}*{\rotatebox{90}{OVRPLTW}} & OR-Tools & 8.728 & 0.656\% & 9.826 & 0.984\% & 14.535 & 1.779\% & 16.263 & 2.287\%  \\ 
    & OR-Tools (x10) & 8.675 & * & 9.710 & * & 14.384 & * & 15.873 & * & & OR-Tools (x10) & 8.669 & * & 9.711 & * & 14.279 & * & 15.884 & *  \\ 
    % & POMO-MTL-pretrain & 9.514 & 9.628\% & 10.288 & 5.710\% & 15.879 & 10.453\% & 16.765 & 5.762\% & & POMO-MTL-pretrain & 8.987 & 3.633\% & 10.347 & 6.376\% & 14.896 & 4.374\% & 16.874 & 6.438\%  \\ 
    % & POMO-MTL-FL & 9.445 & 8.857\% & 10.290 & 5.456\% & 15.732 & 9.450\% & 16.797 & 5.621\% & & POMO-MTL-FL & 8.967 & 3.410\% & 10.348 & 6.096\% & 14.873 & 4.221\% & 16.899 & 6.209\%  \\ 
    & MVMoE-pretrain & 9.486 & 9.308\% &{\underline{10.263}} &{\underline{5.420\%}} & 15.808 & 9.948\% &{\underline{16.704}} &{\underline{5.388\%}} & & MVMoE-pretrain & 8.966 & 3.396\% &{\underline{10.32}} &{\underline{6.077\%}} & 14.828 & 3.903\% &{\underline{16.813}} &{\underline{6.059\%}}  \\ 
    & MVMoE-CPL &{\textbf{9.423}} &{\textbf{8.598\%}} & 10.315 & 5.967\% &{\textbf{15.650}} &{\textbf{8.869\%}} & 16.831 & 6.105\% & & MVMoE-CPL &{\textbf{8.937}} &{\textbf{3.067\%}} & 10.370 & 6.441\% &{\textbf{14.757}} &{\textbf{3.404\%}} & 16.938 & 6.526\%  \\ 
    & MVMoE-MTF &{\underline{9.426}}  &{\underline{8.641\%}} & - & - &{\underline{15.665}}  &{\underline{8.969\%}} & - & - & & MVMoE-MTF & {8.971}  & {3.459\%} & - & - & {14.828}  & {3.899\%} & - & - \\
    & MVMoE-FL & 9.428 & 8.650\% &{\textbf{10.256}} &{\textbf{5.165\%}} & 15.669 & 9.001\% &{\textbf{16.724}} &{\textbf{5.187\%}} & & MVMoE-FL &{\underline{8.951}} &{\underline{3.226\%}} &{\textbf{10.316}} &{\textbf{5.834\%}} &{\underline{14.796}} &{\underline{3.672\%}} &{\textbf{16.841}} &{\textbf{5.853\%}}  \\ 
    \midrule
    \multirow{6}*{\rotatebox{90}{VRPBLTW}} & OR-Tools & 14.890 & 1.402\% & 9.141 & 0.901\% & 25.979 & 2.518\% & 14.991 & 2.205\% & \multirow{6}*{\rotatebox{90}{OVRPBLTW}} & OR-Tools & 8.729 & 0.624\% & 9.826 & 0.987\% & 14.496 & 1.724\% & 16.267 & 2.293\%  \\ 
    & OR-Tools (x10) & 14.677 & * & 9.043 & * & 25.342 & * & 14.655 & * & & OR-Tools (x10) & 8.673 & * & 9.710 & * & 14.250 & * & 15.888 & *  \\ 
    % & POMO-MTL-pretrain & 15.980 & 9.035\% & 9.569 & 5.776\% & 27.247 & 7.746\% & 15.502 & 6.063\% & & POMO-MTL-pretrain & 9.532 & 9.851\% & 10.286 & 5.685\% & 15.738 & 10.498\% & 16.78 & 5.757\%  \\ 
    % & POMO-MTL-FL & 15.881 & 8.354\% & 9.585 & 5.908\% & 26.997 & 6.779\% & 15.535 & 6.238\%  & & POMO-MTL-FL & 9.462 & 9.054\% & 10.291 & 5.451\% & 15.604 & 9.571\% & 16.812 & 5.613\%  \\ 
    & MVMoE-pretrain & 15.945 & 8.775\% &{\underline{9.545}} &{\underline{5.479\%}} & 27.142 & 7.332\% &{\textbf{15.444}} &{\textbf{5.678\%}} & & MVMoE-pretrain & 9.503 & 9.516\% &{\underline{10.261}} &{\underline{5.397\%}} & 15.671 & 10.009\% &{\underline{16.719}} &{\underline{5.381}\%}  \\ 
    & MVMoE-CPL &{\underline{15.842}} &{\underline{8.077\%}} & 9.606 & 6.363\% &{\underline{26.885}} &{\underline{6.326\%}} & 15.595 & 7.083\% & & MVMoE-CPL & 9.451 & 8.925\% & 10.316 & 5.986\% &{\textbf{15.509}} &{\textbf{8.895\%}} & 16.853 & 6.163\%  \\ 
    & MVMoE-MTF &{\textbf{15.832}}  &{\textbf{8.007\%}} & - & - &{\textbf{26.849}}  &{\textbf{6.166\%}} & - & - & & MVMoE-MTF &{\underline{9.446}}  &{\underline{8.879\%}} & - & - &{\underline{15.525}}  &{\underline{8.996\%}} & - & - \\
    & MVMoE-FL & 15.852 & 8.149\% &{\textbf{9.540}} &{\textbf{5.378\%}} & 26.922 & 6.472\% &{\underline{15.457}} &{\underline{5.706\%}} & & MVMoE-FL &{\textbf{9.445}} &{\textbf{8.868\%}} &{\textbf{10.255}} &{\textbf{5.158\%}} & 15.537 & 9.091\% &{\textbf{16.743}} &{\textbf{5.202\%}}  \\ 
    \bottomrule
  \end{tabular}}
%   \end{sc}
  \end{small}
  \end{centering}
  % \vskip -0.15in
\end{table*}

We conduct empirical evaluations to verify the superiority of our proposed MPSF-FL framework. 
% Building upon the work of MVMoE~\cite{zhou2024mvmoe}, and following the our methodology, 
% Our experimental procedure consists of two distinct phases. 
First, we perform global pre-training on 6 simple VRPs, including CVRP, OVRP, VRPB, VRPL, VRPTW, and OVRPTW. Subsequently, we carry out fine-tuning on 10 VRP variants characterized by complex constraints, including OVRPB, OVRPL, VRPBL, VRPBTW, VRPLTW, OVRPBL, OVRPBTW, OVRPLTW, VRPBLTW, and OVRPBLTW. In other words, each client performs fine-tuning on one of the 10 problem variants. In Section \ref{sec:exp_1}, we applied MPSF-FL by taking MVMoE as the global model with its data generation approach~\citep{zhou2024mvmoe}, while in Section \ref{sec:exp_2}, we applied using CaDA with its generated data~\citep{li2025cada}.
All experiments were conducted on a computing platform equipped with Intel(R) Xeon(R) Gold 5120 CPUs @ 2.20GHz and NVIDIA Tesla V100-SXM2-32GB GPUs.

\paragraph{Baselines.} 
We employed PyVRP solver~\citep{wouda2024pyvrp} based on HGS-CVRP~\citep{vidal2022hybrid} and the Google OR-Tools~\citep{ortools_routing} for benchmarking. For problem scales $n=50$ and $n=100$, we set time limits of 20 seconds and 40 seconds, respectively. We also applied 200 and 400 seconds in OR-Tools, denoted as OR-Tools (x10). 
To validate the effectiveness of MPSF-FL, we compared with \textbf{MVMoE-CPL}/\textbf{CaDA-CaDA} that applied CPL on each single VRP of client using the pre-trained global model. 
% with full-finetune cross-problem learning while training on one of the complex problems as the baseline. 
Besides, we also compared a centralized multi-task fine-tuned model \textbf{MVMoE-MTF} to validate the performance of MPSF-FL on trained problem. It was fine-tuned the pre-trained global model on a mixture of datasets from all 10 complex VRP variants, though access to all datasets is often impractical in real-world settings due to data privacy concerns.
% Moreover, we applied  MPSF-FL to \textbf{POMO-MTL} \citep{liu2024multi}, denoted as \textbf{POMO-MTL-FL}. 
All baselines undergo the same pre-training on 6 VRPs.

\paragraph{Training.} During the pre-training phase, we follow all experimental settings and configurations of the unified neural solver used as the global model. For MVMoE, all models undergo 5,000 epochs of global pre-training on a simple task set; while for CaDA, the training epoch is set to 300.
During the fine-tuning phase, for MVMoE, the Adam ptimizer~\citep{Loshchilov2017DecoupledWD} is used with a learning rate of $1e-5$, according to the MVMoE's training parameters in the final stage and avoid overfitting in the training process, a weight decay of $1e-6$, a batch size of 128 and 20,000 instances for each epoch. The training process involves two problem scales, $N \in \{50, 100\}$. 
For MVMoE-CPL, we fine-tuned the pre-trained model with 100 epochs on each complex task, ultimately obtaining 10 models tailored to specific complex tasks respectively. For MVMoE-MTF, we selected the total number of rounds used by all 10 clients in our MPSF-FL, which equals 1,000 ($100 \times 10$). In the fine-tuning phase of MPSF-FL, we set the number of communication rounds to 20, with each client performing 5 local epochs per round. This results in a total of 100 local training epochs, ensuring consistency with the independent fine-tuning baseline. The number of clients corresponds to the number of fine-tuning problems, with each client dedicated to a specific VRP, and full client participation is maintained in each round. For the ties-merging strategy, the parameter mask rate $(100 - \kappa)\%$ is set to $0.8$, and the scaling parameter $\lambda$ is fixed at $1.0$. 

For CaDA, we fine-tune CaDA using the batch size 256 and 100,000 instances for each epoch with a learning rate of $3e-5$ for two problem scales, $N \in \{50, 100\}$. For the CaDA's federated learning process, we maintain most of the above MPSF-FL settings, but with each client performing one local epoch per round, maintaining a training intensity similar to that used during MVMoE training. Consequently, the fine-tuning epoch of CaDA-CPL is set to 20.

% \noindent\textbf{2) MVMoE Configuration:} The MoE layers are configured with $m=4$ experts, activating the top-$k=2$ experts per token. The weight $\alpha$ for the auxiliary load-balancing loss $\mathcal{L}_b$ is set to 0.01. We adopt node-level and input-choice gating for both encoder and decoder layers.
\paragraph{Inference.} 
To evaluate the effectiveness of MPSF-FL in balancing local specialization and global generalization, we analyze the performance of various models on specific VRPs of clients. For every client, we define two key metrics:
\begin{itemize}
    \item \textbf{Trained VRPs Obj.\&Gap:} The performance evaluated on the VRP that a neural solver was fine-tuned on.
    \item \textbf{Unseen VRPs Obj.\&Gap:} The average performance evaluated on the other 9 complex VRPs that the neural solver was not fine-tuned. For MVMoE-MTF, it is fine-tuned on all VRPs, so there are no unseen problems, which is indicated by ``-'' in Table \ref{mvmoe_client}.
\end{itemize}

For all neural solvers, we use greedy rollout with $\times8$ augmentation following POMO~\citep{kwon2020pomo}. We report average results over 1K instances per VRP. Gaps are gauged against the best results from traditional VRP solvers.

\begin{table*}[!t]
  % \vskip -0.05in
  \caption{Client-side adaptation performance comparison for Cada on 1K test instances. The best performance is highlighted in bold, and the second-best is underlined.}
  \label{cada_client}
  % \vskip -0.1in
  \begin{centering}
  \begin{small}
  \renewcommand\arraystretch{1.2}  % 0.95
  \resizebox{1.0\textwidth}{!}{ 
  \begin{tabular}{ll|cccccccc|ll|cccccccc}
    \toprule
    \multicolumn{2}{c|}{\multirow{2}{*}{Method}} & \multicolumn{4}{c}{\textbf{$n=50$}} & \multicolumn{4}{c|}{$n=100$} & \multicolumn{2}{c|}{\multirow{2}{*}{Method}} &
    \multicolumn{4}{c}{\textbf{$n=50$}} & \multicolumn{4}{c}{$n=100$} \\
    \cmidrule(lr){3-6} \cmidrule(lr){7-10} \cmidrule(lr){13-16} \cmidrule(lr){17-20}
     & & Trained Obj. & Trained Gap & Unseen Obj. & Unseen Gap & Trained Obj. & Trained Gap & Unseen Obj. & Unseen Gap & & & Trained Obj. & Trained Gap & Unseen Obj. & Unseen Gap & Trained Obj. & Trained Gap & Unseen Obj. & Unseen Gap \\
     \midrule
        \multirow{5}*{\rotatebox{90}{OVRPB}} & HGS-PyVRP & 6.898 & * & 12.297 & * & 10.335 & * & 19.457 & *  & \multirow{5}*{\rotatebox{90}{OVRPL}} & HGS-PyVRP & 6.507 & * & 12.341 & * & 9.724 & * & 19.525 & *  \\ 
        & OR-Tools & 6.928 & 0.412\% & 12.322 & 0.418\% & 10.577 & 2.315\% & 19.742 & 2.001\%  & & OR-Tools & 6.552 & 0.668\% & 12.364 & 0.389\% & 10.001 & 2.791\% & 19.806 & 1.948\%  \\ 
        & CaDA-pretrained & 7.660 & 10.953\% & 12.921 & 5.885\% & 11.237 & 8.659\% &{\underline{20.793}} &{\underline{6.886\%}} & & CaDA-pretrained & 7.497 & 15.054\% &{\underline{12.940}} &{\underline{5.429\%}} & 10.365 & 6.621\% & 20.890 & 7.113\% \\
        & CaDA-CPL &{\textbf{7.049}} &{\textbf{2.152\%}} &{\underline{12.871}} &{\underline{5.206\%}} &{\textbf{10.750}} &{\textbf{3.988\%}} & 20.912 & 6.943\% & & CaDA-CPL &{\underline{6.662}} &{\underline{2.353\%}} & 12.949 & 5.452\% &{\textbf{10.095}} &{\textbf{3.776\%}} &{\underline{20.886}} & {\underline{6.834\%}} \\
        % & CaDA-MTF & 7.045 & 2.094\% & - & - & 10.746 & 3.949\% & - & - & & CaDA-MTF & 6.658 & 2.286\% & - & - & 10.103 & 3.860\% & - & - \\
        & CaDA-FL &{\underline{7.053}} &{\underline{2.213\%}} &{\textbf{12.532}} &{\textbf{1.950\%}} &{\underline{10.767}} &{\underline{4.148\%}} &{\textbf{20.041}} &{\textbf{3.148\%}} & & CaDA-FL &{\textbf{6.656}} &{\textbf{2.255\%}} &{\textbf{12.576}} &{\textbf{1.938\%}} &{\underline{10.109}} &{\underline{3.923\%}} &{\textbf{20.126}} &{\textbf{3.271\%}} \\
        \midrule
        \multirow{5}*{\rotatebox{90}{VRPBL}} & HGS-PyVRP & 10.186 & * & 11.932 & * & 14.779 & * &  18.963 & *  & \multirow{5}*{\rotatebox{90}{VRPBTW}} & HGS-PyVRP & 18.292 & * & 11.031 & * & 29.467 & * & 17.331 & * \\ 
        & OR-Tools & 10.331 & 1.390\% & 11.944 & 0.309\% & 15.426 & 4.338\% & 19.203 & 1.776\% & & OR-Tools & 18.366 & 0.383\% & 11.051 & 0.421\% & 29.945 & 1.597\% & 17.590 & 2.081\%  \\ 
        & CaDA-pretrained & 10.704 & 4.948\% & 12.583 & 6.552\% & 15.625 & 5.644\% &{\underline{20.306}} &{\underline{7.221\%}} & & CaDA-pretrained & 19.165 & 4.751\% & 11.643 & 6.574\% & 31.909 & 8.268\% &{\underline{18.497}} &{\underline{6.930\%}} \\
        & CaDA-CPL &{\textbf{10.532}} &{\textbf{3.342\%}} &{\underline{12.573}} &{\underline{6.238\%}} &{\textbf{15.488}} &{\textbf{4.758\%}} & 20.403 & 7.975\% & & CaDA-CPL &{\underline{18.526}} &{\underline{1.265\%}} &{\underline{11.565}} &{\underline{6.148\%}} & 30.161 & 2.346\% & 18.342 & 6.988\% \\
        % & CaDA-MTF & 10.530 & 3.322\% & - & - & 15.487 & 4.748\% & - & - & & CaDA-MTF & 18.496 & 1.100\% & - & - & 30.087 & 2.090\% & - & - \\
        & CaDA-FL &{\underline{10.542}} &{\underline{3.438\%}} &{\textbf{12.130}} &{\textbf{1.765\%}} &{\underline{15.509}} &{\underline{4.897\%}} &{\textbf{19.524}} &{\textbf{3.213\%}} & & CaDA-FL &{\textbf{18.526}} &{\textbf{1.264\%}} &{\textbf{11.252}} &{\textbf{2.041\%}} &{\textbf{30.161}} &{\textbf{2.345\%}} &{\textbf{17.875}} &{\textbf{3.335\%}} \\
        \midrule
        \multirow{5}*{\rotatebox{90}{VRPLTW}} & HGS-PyVRP & 16.356 & * & 11.247 & * & 25.757 & * & 17.743 & * & \multirow{5}*{\rotatebox{90}{OVRPBL}} & HGS-PyVRP & 6.899 & * & 12.297 & * & 10.335 & * & 19.457 & * \\ 
        & OR-Tools & 16.441 & 0.499\% & 11.265 & 0.408\% & 26.259 & 1.899\% & 17.999 & 2.047\% & & OR-Tools & 6.927 & 0.386\% & 12.322 & 0.421\% & 10.582 & 2.363\% & 19.741 & 1.996\% \\ 
        & CaDA-pretrained & 16.772 & 2.471\% &{\underline{11.909}} &{\underline{6.827\%}} & 26.749 & 3.785\% &{\underline{19.070}} &{\underline{7.428\%}} & & CaDA-pretrained & 7.709 & 11.641\% & 12.916 & 5.809\% & 11.568 & 11.834\% &{\underline{20.757}} &{\underline{6.533\%}} \\
        & CaDA-CPL &{\textbf{16.669}} &{\textbf{1.892\%}} & 11.923 & 7.231\% &{\textbf{26.560}} &{\textbf{3.075\%}} & 19.128 & 8.021\% & & CaDA-CPL &{\textbf{7.052}} &{\textbf{2.190\%}} &{\underline{12.804}} &{\underline{3.899\%}} &{\textbf{10.754}} &{\textbf{4.024\%}} & 21.030 & 7.365\% \\
        % & CaDA-MTF & 16.657 & 1.811\% & - & - & 26.517 & 2.910\% & - & - & & CaDA-MTF & 7.047 & 2.112\% & - & - & 10.749 & 3.976\% & - & - \\
        & CaDA-FL &{\underline{16.678}} &{\underline{1.935\%}} &{\textbf{11.456}} &{\textbf{1.948\%}} &{\underline{26.572}} &{\underline{3.129\%}} &{\textbf{18.287}} &{\textbf{3.289\%}} & & CaDA-FL &{\underline{7.061}} &{\underline{2.315\%}} &{\textbf{12.539}} &{\textbf{2.022\%}} &{\underline{10.766}} &{\underline{4.142\%}} &{\textbf{20.057}} &{\textbf{3.234\%}} \\
        \midrule
        \multirow{5}*{\rotatebox{90}{OVRPBTW}} & HGS-PyVRP & 11.669 & * & 11.767 & * & 19.156 & * & 19.156 & *  & \multirow{5}*{\rotatebox{90}{OVRPLTW}} & HGS-PyVRP & 10.510 & * & 11.896 & * & 16.926 & * & 18.725 & * \\ 
        & OR-Tools & 11.682 & 0.109\% & 11.794 & 0.451\% & 19.303 & 0.757\% & 18.772 & 2.174\%, & & OR-Tools       & 10.497 & -0.114\% & 11.926 & 0.476\% & 17.023 & 0.728\% & 19.025 & 2.177\% \\ 
        & CaDA-pretrained & 12.145 & 4.026\% & 12.423 & 6.655\% & 20.556 & 7.235\% & 19.758 & 7.044\% & & CaDA-pretrained & 10.620 & 1.034\% & 12.593 & 6.987\% & 17.326 & 2.341\% & 20.117 & 7.588\% \\
        & CaDA-CPL &{\underline{11.772}} &{\underline{0.867\%}} &{\underline{12.340}} &{\underline{6.344\%}} &{\underline{19.495}} &{\underline{1.751\%}} &{\underline{19.325}} &{\underline{5.517\%}} & & CaDA-CPL &{\underline{10.616}} &{\underline{0.994\%}} &{\underline{12.561}} &{\underline{6.361\%}} &{\textbf{17.232}} &{\textbf{1.785\%}} &{\underline{20.111}} &{\underline{7.273\%}} \\
        % & CaDA-MTF & 11.757 & 0.748\% & - & - & 19.443 & 1.478\% & - & - & & CaDA-MTF & 10.608 & 0.909\% & - & - & 17.223 & 1.730\% & - & - \\
        & CaDA-FL &{\textbf{11.769}} &{\textbf{0.848\%}} &{\textbf{12.006}} &{\textbf{2.089\%}} &{\textbf{19.494}} &{\textbf{1.745\%}} &{\textbf{19.061}} &{\textbf{3.396\%}} & & CaDA-FL &{\textbf{10.616}} &{\textbf{0.982\%}} &{\textbf{12.153}} &{\textbf{2.192\%}} &{\underline{17.239}} &{\underline{1.827\%}} &{\textbf{19.336}} &{\textbf{3.467\%}} \\
        \midrule
        \multirow{5}*{\rotatebox{90}{VRPBLTW}} & HGS-PyVRP & 18.361 & * & 10.998 & * & 29.026 & * & 17.293 & * & \multirow{5}*{\rotatebox{90}{OVRPBLTW}} & HGS-PyVRP & 11.668 & * & 11.767 & * & 19.156 & * & 18.477 & * \\ 
        & OR-Tools & 18.422 & 0.332\% & 11.045 & 0.427\% & 29.830 & 2.770\% & 17.602 & 1.951\% & & OR-Tools & 11.681 & 0.106\% & 11.794 & 0.452\% & 19.305 & 0.767\% & 18.772 & 2.173\% \\ 
        & CaDA-pretrained & 19.554 & 5.163\% &{\underline{11.600}} &{\underline{6.528\%}} & 32.450 & 8.812\% &{\underline{18.436}} &{\underline{6.869\%}} & & CaDA-pretrained & 12.127 & 3.878\% & 12.425 & 6.671\% & 20.591 & 7.436\% & 19.754 & 7.022\% \\
        & CaDA-CPL &{\underline{18.878}} &{\underline{1.540\%}} & 11.566 & 6.687\% &{\textbf{30.627}} &{\textbf{2.719\%}} & 18.468 & 8.259\% & & CaDA-CPL &{\textbf{11.770}} &{\textbf{0.860\%}} &{\underline{12.246}} &{\underline{5.226\%}} &{\underline{19.497}} &{\underline{1.756\%}} &{\underline{19.297}} &{\underline{5.151\%}} \\
        % & CaDA-MTF & 18.839 & 1.326\% & - & - & 30.537 & 2.409\% & - & - & & CaDA-MTF & 11.758 & 0.757\% & - & - & 19.441 & 1.470\% & - & - \\
        & CaDA-FL &{\textbf{18.870}} &{\textbf{1.494\%}} &{\textbf{11.201}} &{\textbf{1.929\%}} &{\underline{30.633}} &{\underline{2.733\%}} &{\textbf{17.826}} &{\textbf{3.324\%}} & & CaDA-FL &{\underline{11.771}} &{\underline{0.867\%}} &{\textbf{12.013}} &{\textbf{2.152\%}} &{\textbf{19.494}} &{\textbf{1.745\%}} &{\textbf{19.061}} &{\textbf{3.382\%}} \\
     \bottomrule
  \end{tabular}}
%   \end{sc}
  \end{small}
  \end{centering}
  % \vskip -0.15in
\end{table*}

\begin{table*}[!t]
  % \vskip -0.05in
  \caption{Client-side performance comparison in data-scarce scenario.}
  \label{data_scarcity}
  % \vskip -0.1in
  \begin{centering}
  \begin{small}
  \renewcommand\arraystretch{1.2}  % 0.95
  \resizebox{1.0\textwidth}{!}{ 
  \begin{tabular}{ll|cccccccc|ll|cccccccc}
    \toprule
    \multicolumn{2}{c|}{\multirow{2}{*}{Method}} & \multicolumn{4}{c}{\textbf{$n=50$}} & \multicolumn{4}{c|}{$n=100$} & \multicolumn{2}{c|}{\multirow{2}{*}{Method}} &
    \multicolumn{4}{c}{\textbf{$n=50$}} & \multicolumn{4}{c}{$n=100$} \\
    \cmidrule(lr){3-6} \cmidrule(lr){7-10} \cmidrule(lr){13-16} \cmidrule(lr){17-20}
     & & Trained Obj. & Trained Gap & Unseen Obj. & Unseen Gap & Trained Obj. & Trained Gap & Unseen Obj. & Unseen Gap & & & Trained Obj. & Trained Gap & Unseen Obj. & Unseen Gap & Trained Obj. & Trained Gap & Unseen Obj. & Unseen Gap \\
     \midrule
     \multirow{4}*{\rotatebox{90}{OVRPB}} & MVMoE-CPL & 6.018 & 4.719\% & 10.751 & 6.670\% & \textbf{8.783} & \textbf{4.986\%} & 17.533 & 5.969\% & \multirow{4}*{\rotatebox{90}{OVRPL}} & MVMoE-CPL & 6.647 & 2.408\% & 10.674 & 6.983\% & \textbf{10.065} & \textbf{2.820\%} & 17.373 & 6.251\% \\
     & MVMoE-FL   & \textbf{6.011} & \textbf{4.596\%} & \textbf{10.623} & \textbf{5.527\%} & 8.804 & 5.243\% & \textbf{17.467} & \textbf{5.547\%} & & MVMoE-FL   & \textbf{6.636} & \textbf{2.237\%} & \textbf{10.564} & \textbf{5.908\%} & 10.071 & 2.881\% & \textbf{17.340} & \textbf{5.919\%} \\
     \cmidrule(lr){2-2} \cmidrule(lr){3-10} \cmidrule(lr){12-12} \cmidrule(lr){13-20}
     & CaDA-CPL & \textbf{7.052} & \textbf{2.205\%} & 12.869 & 5.204\% & \textbf{10.764} & \textbf{4.119\%} & 20.823 & 6.613\% & & CaDA-CPL & \textbf{6.660} & \textbf{2.318\%} & 12.938 & 5.373\% & \textbf{10.100} & \textbf{3.829\%} & 20.885 & 6.882\% \\
     & CaDA-FL & 7.053 & 2.213\% & \textbf{12.528} & \textbf{1.931\%} & 10.772 & 4.198\% & \textbf{20.058} & \textbf{3.246\%} & & CaDA-FL & 6.665 & 2.397\% & \textbf{12.574} & \textbf{1.945\%} & 10.111 & 3.942\% & \textbf{20.136} & \textbf{3.323\%} \\
      \midrule
     \multirow{4}*{\rotatebox{90}{VRPBL}} & MVMoE-CPL & 8.178 & 1.865\% & 10.569 & 7.937\% & \textbf{11.947} & \textbf{1.362\%} & 17.182 & 6.685\% & \multirow{4}*{\rotatebox{90}{VRPBTW}} & MVMoE-CPL & 15.935 & 8.049\% & 9.674 & 7.481\% & \textbf{27.015} & \textbf{6.219\%} & 15.583 & 7.081\% \\
     & MVMoE-FL & \textbf{8.174} & \textbf{1.809\%} & \textbf{10.410} & \textbf{6.191\%} & 11.952 & 1.403\% & \textbf{17.158} & \textbf{6.421\%} & & MVMoE-FL & \textbf{15.929} & \textbf{7.974\%} & \textbf{9.531} & \textbf{5.383\%} & 27.025 & 6.267\% & \textbf{15.443} & \textbf{5.706\%} \\
     \cmidrule(lr){2-2} \cmidrule(lr){3-10} \cmidrule(lr){12-12} \cmidrule(lr){13-20}
     & CaDA-CPL & \textbf{10.539} & \textbf{3.399\%} & 12.557 & 6.064\% & \textbf{15.502} & \textbf{4.841\%} & 20.411 & 8.416\% & & CaDA-CPL & 18.554 & 1.410\% & 11.546 & 6.048\% & 30.246 & 2.636\% & 18.341 & 6.974\% \\
     & CaDA-FL & 10.546 & 3.470\% & \textbf{12.131} & \textbf{1.764\%} & 15.521 & 4.964\% & \textbf{19.528} & \textbf{3.242\%} & & CaDA-FL & \textbf{18.527} & \textbf{1.267\%} & \textbf{11.247} & \textbf{2.004\%} & \textbf{30.178} & \textbf{2.402\%} & \textbf{17.889} & \textbf{3.382\%} \\
     \midrule
     \multirow{4}*{\rotatebox{90}{VRPLTW}} & MVMoE-CPL & \textbf{14.924} & \textbf{2.339\%} & 9.763 & 7.677\% & \textbf{25.500} & \textbf{1.416\%} & 15.744 & 7.277\% & \multirow{4}*{\rotatebox{90}{OVRPBL}} & MVMoE-CPL & 6.008 & 4.652\% & 10.760 & 6.734\% & \textbf{8.772} & \textbf{5.077\%} & 17.536 & 5.966\%\\
     & MVMoE-FL & 14.942 & 2.460\% & \textbf{9.630} & \textbf{5.869\%} & 25.527 & 1.529\% & \textbf{15.603} & \textbf{6.135\%} & & MVMoE-FL & \textbf{6.000} & \textbf{4.521\%} & \textbf{10.623} & \textbf{5.524\%} & 8.783 & 5.216\% & \textbf{17.468} & \textbf{5.535\%} \\
     \cmidrule(lr){2-2} \cmidrule(lr){3-10} \cmidrule(lr){12-12} \cmidrule(lr){13-20}
     & CaDA-CPL & \textbf{16.664} & \textbf{1.849\%} & 11.926 & 7.208\% & 26.581 & 3.158\% & 19.271 & 8.897\% & & CaDA-CPL & \textbf{7.054} & \textbf{2.206\%} & 12.814  & 4.044\% & \textbf{10.764} & \textbf{4.121\%} & 20.920 & 6.856\% \\
     & CaDA-FL & 16.674 & 1.917\% & \textbf{11.458} & \textbf{1.979\%} & \textbf{26.574} & \textbf{3.137\%} & \textbf{18.296} & \textbf{3.345\%} & & CaDA-FL & 7.054 & 2.213\% & \textbf{12.535} & \textbf{1.987\%} & 10.771 & 4.188\% & \textbf{20.060} & \textbf{3.244\%} \\
      \midrule
     \multirow{4}*{\rotatebox{90}{OVRPBTW}} & MVMoE-CPL & 9.442 & 8.810\% & 10.335 & 6.338\% & 15.686 & 9.111\% & 16.816 & 6.069\% & \multirow{4}*{\rotatebox{90}{OVRPLTW}} & MVMoE-CPL & \textbf{8.946} & \textbf{3.165\%} & 10.417 & 7.119\% & \textbf{14.775} & \textbf{3.534\%} & 16.941 & 6.607\% \\
     & MVMoE-FL & \textbf{9.434} & \textbf{8.727\%} & \textbf{10.252} & \textbf{5.143\%} & \textbf{15.678} & \textbf{9.064\%} & \textbf{16.714} & \textbf{5.165\%} & &  MVMoE-FL & 8.953 & 3.248\% & \textbf{10.313} & \textbf{5.809\%} & 14.802 & 3.706\% & \textbf{16.827} & \textbf{5.807\%} \\
     \cmidrule(lr){2-2} \cmidrule(lr){3-10} \cmidrule(lr){12-12} \cmidrule(lr){13-20}
     & CaDA-CPL & 11.781 & 0.952\% & 12.301 & 6.073\% &  19.546 & 2.012\% & 19.392 & 5.943\% & & CaDA-CPL & \textbf{10.612} & \textbf{0.953\%} & 12.577 & 6.366\% & \textbf{17.238} & \textbf{1.816\%} & 20.190 & 7.660\%\\
     & CaDA-FL & \textbf{11.771} & \textbf{0.862\%} & \textbf{12.012} & \textbf{2.131\%} & \textbf{19.500} & \textbf{1.777\%} & \textbf{19.080} & \textbf{3.498\%} & & CaDA-FL & 10.614 & 0.971\% & \textbf{12.158} & \textbf{2.218\%} & 19.547 & 2.020\% & \textbf{19.094} & \textbf{3.518\%} \\
     \midrule
     \multirow{4}*{\rotatebox{90}{VRPBLTW}} & MVMoE-CPL & 15.864 & 8.250\% & 9.672 & 7.320\% & 26.946 & 6.567\% & 15.585 & 6.939\%  & \multirow{4}*{\rotatebox{90}{OVRPBLTW}} & MVMoE-CPL & 9.451 & 8.932\% & 10.333 & 6.321\% & 15.542 & 9.123\% & 16.831 & 6.027\% \\
     & MVMoE-FL & \textbf{15.858} & \textbf{8.184\%} & \textbf{9.539} & \textbf{5.355\%} & \textbf{26.935} & \textbf{6.530\%} & \textbf{15.453} & \textbf{5.677\%} & & MVMoE-FL & \textbf{9.448} & \textbf{8.903\%} & \textbf{10.252} & \textbf{5.138\%} & \textbf{15.542} & \textbf{9.122\%} & \textbf{16.726} & \textbf{5.140\%} \\
     \cmidrule(lr){2-2} \cmidrule(lr){3-10} \cmidrule(lr){12-12} \cmidrule(lr){13-20}
     & CaDA-CPL & 18.906 & 1.683\% & 11.517 & 6.194\% & 30.759 & 3.151\% & 18.307 & 7.146\% & & CaDA-CPL & 11.783 & 0.959\% & 12.229 & 5.075\% & 19.547 & 2.013\% & 19.331 & 5.358\% \\
     & CaDA-FL & \textbf{18.878} & \textbf{1.536\%} & \textbf{11.207} & \textbf{1.989\%} & \textbf{30.659} & \textbf{2.820\%} & \textbf{17.832} & \textbf{3.345\%} & & CaDA-FL & \textbf{11.772} & \textbf{0.876\%} & \textbf{12.005} & \textbf{2.092\%} & \textbf{19.503} & \textbf{1.789\%} & \textbf{19.082} & \textbf{3.485\%} \\
    \bottomrule
  \end{tabular}}
%   \end{sc}
  \end{small}
  \end{centering}
  % \vskip -0.15in
\end{table*}

\subsection{Performance Comparison}
\label{sec:exp_1}

% Table \ref{mvmoe_client} summarizes the results for clients. Our analysis yields several critical insights:
In this section, we present the results of applying MPSF-FL to MVMoE. We denote the client model of MPSF-FL as \textbf{MVMoE-FL}. As shown in Table \ref{mvmoe_client}, our MVMoE-FL outperformed MVMoE-CPL on all unseen tasks, demonstrating its significant enhancement of the local client's generalization capability for unseen problems. Although MVMoE-FL exhibits inferior performance to the MVMoE-CPL which is task-specialized on the trained problem, it demonstrates problem-solving capabilities comparable to the multi-task fine-tuned model MVMoE-MTF on trained problems.
Across a total of 20 problem scenarios, MPSF-FL performed comparably to or even better than MVMoE-MTF in 12 cases. A notable example emerges in VRPLTW ($N=100$), where MVMoE-FL achieves a trained gap of 1.464\% and an unseen gap of 6.098\%. Compared to MVMoE-MTF's 1.526\% trained gap, this represents a 0.062\% improvement, and in comparison to MVMoE-CPL's 7.168\% unseen gap, it achieves a 1.07\% improvement. 

The above experimental results demonstrate that our MPSF-FL enables local clients to obtain models with strong multi-task generalization capabilities while achieving performance comparable to the multi-task fine-tuning method, while significantly safeguarding data privacy. 
% Additionally, our framework has also obtained a global model that integrates the solving capabilities across all problems, with its performance evaluated in Appendix \ref{app:server}.

% the VRPBLTW ($N=100$), where MVMoE-FL achieves an Unseen Gap of $5.731\%$, significantly outperforming MVMoE-CPL, which lags behind at $7.083\%$. The difference is as high as $1.3\%$. Similarly, in the VRPBTW ($N=100$) task, MVMoE-FL maintains a lower Unseen Gap $5.731\%$ compared to MVMoE-CPL's $7.056\%$. Moreover, their training gap differs by a mere $0.15\%$, which suggests that the federated mechanism can enhance the generalization capabilities of local models while allowing for task-specific adaptation. Even on simpler variants like OVRPB ($N=50$), MVMoE-FL retains its advantage with an Unseen Gap of $5.528\%$ versus $5.833\%$ for the CPL baseline. These results collectively validate that MVMoE-FL does not merely compromise performance for privacy; rather, it leverages client-side diversity to escape local optima, yielding a more generalizable solver than traditional CPL paradigm.

\subsection{Versatility}
\label{sec:exp_2}
Our MPSF-FL is versatile to differet unified models for VRPs. We further deploy it with CaDA~\citep{li2025cada}, which achieves state-of-the-art performance in solving multi-task VRPs. Table \ref{cada_client} details the performance of the CaDA model under different learning paradigms. Across all VRP variants and problem scales, CaDA-FL consistently achieves a significantly lower optimality gap on unseen instances compared to CaDA-CPL. For instance, in VRPBL ($N=50$),  CaDA-FL reduces the unseen gap from CaDA-CPL's 6.238\% to 1.765\%, and in VRPLTW ($N=100$), it drops from 8.021\% to 3.289\%. This demonstrates that the MPSF-FL framework possesses sufficient versatility for deployments with different unified neural solvers. Moreover, CaDA-FL also excels in training adaptation, outperforming CaDA-CPL on the trained gap metric in multiple scenarios. Specifically, CaDA-FL achieves lower trained gaps in OVRPL ($N=50$), VRPBLTW ($N=50$) and OVRPBTW (both $N=50$ and $N=100$). This evidence suggests that for complex constraint-aware architectures like CaDA, the collaborative learning mechanism of FL is capable of yielding a solver that is not only more robust but also more precise in handling known constraints. 
% In Appendix \ref{app:benchmark_performance}, we also compare CaDA-CPL and CaDA-FL on VRPBTW benchmark instances as used in \citep{gelinas1995new,reimann2002insertion} to further show the advantage of our MPSF-FL framework.

\subsection{Robustness to Data Scarcity}
In many real-world scenarios, edge clients have access to only a limited number of training instances or employ only limited data due to business privacy consideration. To investigate whether the benefits of federated learning persist under such practical constraints, we conduct experiments in a data scarcity situation where each client is restricted to 20,000 training instances. While this volume might allow for basic convergence, it is significantly lower than the millions of samples typically required for full-scale VRP solver training. We compare \textbf{MVMoE-FL}/\textbf{CaDA-FL} against \textbf{MVMoE-CPL}/\textbf{CaDA-CPL} baseline, respectively, to evaluate the effectiveness of MTSF-FL framework.

Table \ref{data_scarcity} presents the results under the  data scarcity setting. Our MTSF-FL framework maintains obvious advantage. It consistently outperforms the CPL baseline on unseen generalization across all problem variants. More importantly, we observe a significant performance degradation in MVMoE-CPL on many trained problems with $N=50$, which fails to fit the training distribution as effectively as MVMoE-FL. This phenomenon suggests that in data scarcity setting, traditional CPL paradigm may carry the risk of getting stuck in local optima. In contrast, the collaborative aggregation in MVMoE-FL acts as an effective regularizer, enabling the client model to learn more generalizable common knowledge even from limited static data.

\section{Conclusion and Future Work}

We present a ``Multi-problem Pre-train, then Single-problem Fine-tune'' framework with Federated Learning, applying federated learning to address the deficiencies of traditional CPL paradigm in knowledge transfer among complex constraints. We conducted experiments on 10 complex VRP variants using state-of-the-art multi-task solvers and performed comparative evaluations under simulated data-scarcity conditions typical of real-world applications. Results show that our framework successfully enhances CPL ability to generalize to unseen variants, demonstrating excellent versatility and practicality. Our future work will optimize the federated learning framework by conducting in-depth analysis of model parameter distribution changes among different problems. We also intend to leverage more appropriate model merging strategy for enhancing client-side performance on trained problems and generating powerful global model in the central serve.
% we will focus on leveraging more appropriate federated learning strategy for enhancing client-side performance on trained problem.

\newpage
\bibliographystyle{named}
\bibliography{main}

%%%%%%%%%%%%%%%%%%%%%%%%%%%%%%%%%%%%%%%%%%%%%%%%%%%%%%%%%%%%%%%%%%%%%%%%%%%%%%%
%%%%%%%%%%%%%%%%%%%%%%%%%%%%%%%%%%%%%%%%%%%%%%%%%%%%%%%%%%%%%%%%%%%%%%%%%%%%%%%
% APPENDIX
%%%%%%%%%%%%%%%%%%%%%%%%%%%%%%%%%%%%%%%%%%%%%%%%%%%%%%%%%%%%%%%%%%%%%%%%%%%%%%%
%%%%%%%%%%%%%%%%%%%%%%%%%%%%%%%%%%%%%%%%%%%%%%%%%%%%%%%%%%%%%%%%%%%%%%%%%%%%%%%
% \newpage
% \appendix
% \twocolumn

% \input{chapter/Appendix}

\end{document}